# A.N.N.C.R.I.P.S – Artificial Neural Networks for Cancer Research In Prediction & Survival

Amit Mathapati

Cornell University

May 2009

amitmathapati@gmail.com
arm259@cornell.edu

*Abstract*— Prostate cancer stands as the most frequently diagnosed cancer among men aged 50 and older. Contemporary diagnostic and screening procedures predominantly rely on blood tests to assess prostate-specific antigen (PSA) levels and Digital Rectal Examinations (DRE). Regrettably, these methods are plagued by a substantial occurrence of false-positive results (FPTRs), which can engender unwarranted anxiety and invasive follow-up procedures for patients. To address these pressing issues, this research project seeks to harness the potential of intelligent Artificial Neural Networks (ANNs).

This study's overarching objective is to develop an advanced mathematical model specifically tailored to enhance the early detection of prostate cancer, thus facilitating prompt medical intervention and ultimately improving patient outcomes. By seamlessly integrating ANNs into the diagnostic process, we aim to enhance both the accuracy and reliability of prostate cancer screening, thereby drastically reducing the incidence of FPTRs. This model signifies a promising solution for healthcare practitioners to furnish more precise and timely assessments of their patients' conditions.

In the pursuit of these objectives, we will meticulously execute a series of rigorous testing and validation procedures, coupled with comprehensive training of the ANN model, using meticulously curated and diverse datasets. The ultimate aspiration is to create a deployable and marketable solution grounded in this mathematical model, seamlessly adaptable to various healthcare settings, including screening centers, hospitals, and research institutions. This innovative approach has the potential to revolutionize prostate cancer screening, contributing to elevated standards of patient care and early intervention, with the goal of saving lives and mitigating the substantial burden imposed by this prevalent disease.

## I. Introduction

### A. The Genesis & Emphasis on Prostate Cancer?

Prostate cancer represents the foremost prevalent form of non-cutaneous malignancy among the male population in the United States, as substantiated by authoritative sources[1]. This insidious disease casts its shadow over the lives of an alarming one in every six men, underscoring its substantial public health impact. Intriguingly, it is worth noting that a non-smoking male individual is at a notably heightened risk of receiving a prostate cancer diagnosis when juxtaposed with the cumulative risk of all other cancer types combined. This staggering fact underscores the paramount importance of addressing prostate cancer as a top-tier healthcare concern.

Moreover, the relative incidence of prostate cancer diagnosis eclipses that of breast cancer among women, further underscoring the gravity of the issue at hand. In contemporary times, the prevalence of this disease has surged to a staggering extent, with a conservative estimate suggesting that a strikingly high number of over two million men in the United States are currently grappling with the complex challenges posed by prostate cancer.

It is imperative to acknowledge that prostate cancer does not discriminate based on age, as all men stand vulnerable to its insidious onset. However, a notable correlation exists between the incidence of this disease and advancing age, as well as a pertinent familial predisposition. The inexorable passage of time manifests as a precipitating factor, increasing the likelihood of an individual's susceptibility to prostate cancer. Hence, the nexus between age and the diagnosis of prostate cancer merits substantial attention.

The year 2009 stands as a pivotal milestone in the narrative of prostate cancer epidemiology. During this significant period, statistical projections cast a sobering light on the disease's prevalence, with an estimated 192,000 men anticipated to receive a prostate cancer diagnosis. Tragically, this affliction exacted an even graver toll, with over 27,000 men succumbing to its relentless progression. These statistics serve as a stark reminder of the pressing need to channel resources and research endeavors towards combating prostate cancer's profound public health implications.

### B. Current Screening Methods: A Comprehensive Overview

At present, the landscape of prostate cancer diagnosis predominantly relies upon two principal methodologies: the assessment of Protein Specific Antigen (PSA) levels in the bloodstream and the undertaking of Digital Rectum Examination (DRE). These modalities hold a pervasive presence across the spectrum of medical institutions and research establishments, serving as the primary means to detect and evaluate the presence of prostate cancer. However, it is imperative to recognize that they are not without their inherent shortcomings, chief among them being the propensity to yield a substantial number of False Positive Test Results (FPTRs).

PSA, a protein produced by the prostate gland in nominal quantities, takes center stage as a linchpin in prostate cancer diagnosis. When prostate-related issues arise, the production and release of PSA can escalate at an alarming rate, propagating into various parts of the body via the circulatory system. Notably, PSA levels are categorized into three distinct ranges for diagnostic purposes: levels below 4 nanograms per milliliter (ng/mL) are generally deemed within the normal range. A reading between 4 to 10 ng/mL is classified as an intermediate level, while PSA levels soaring above the 10 ng/mL threshold are ominously associated with a heightened risk of prostate cancer among patients.

In parallel, the Digital Rectum Examination (DRE) constitutes a tangible approach to appraising the prostate's physical state. However, this assessment method often invokes skepticism and apprehension among patients, as it necessitates a palpation-based examination of the prostate to detect any irregularities in its shape, texture, or overall formation. At the DRE stage, the attending physician can, to a certain extent, discern the likelihood of prostate cancer or other male-specific malignancies. However, the DRE, though informative, predominantly serves as a preliminary indicator rather than a definitive diagnostic tool. It frequently guides medical practitioners towards the need for further, more invasive procedures, such as biopsies.

Biopsies, though indispensable for securing a conclusive diagnosis, present their own set of challenges. This invasive procedure entails the insertion of a needle into the prostate gland to procure tissue samples, guided by the aid of ultrasound imaging. Regrettably, the biopsy process is notably painful and discomforting, often dissuading a significant number of patients from undergoing the procedure. Consequently, a substantial cohort of potential prostate cancer cases remains undiagnosed due to this aversion to invasive testing, exacerbating the diagnostic conundrum.

It is crucial to underscore that despite the widespread utilization of the aforementioned methodologies, the specter of FPTRs looms large. False positive results not only trigger undue psychological distress for patients but also generate a cascade of unnecessary follow-up procedures. To address this vexing issue, the present research endeavors to harness the potency of intelligent Artificial Neural Networks (ANNs) to construct an adept, readily deployable, and marketable mathematical model. This model aims to circumvent the need for a protracted series of trial-and-error methods, ultimately expediting the diagnosis and initiation of treatment at an earlier juncture in the disease progression. Through the strategic implementation of ANNs, this research aspires to revolutionize the landscape of prostate cancer detection, mitigating the deleterious consequences of FPTRs, and ushering in a new era of precise, timely, and patient-friendly diagnostics and therapeutics.

## II. ANNCRIPS

### A. Artificial Neural Networks

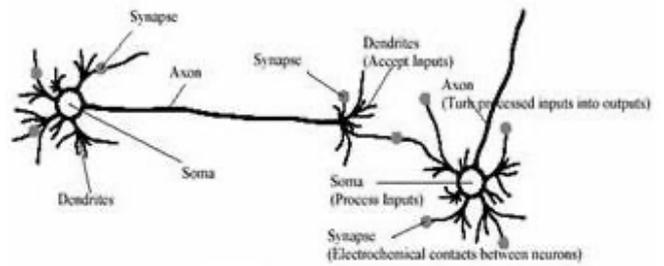

Fig. 1 Neuron model in human brain

Artificial Neural Networks (ANNs) represent a profoundly intriguing and innovative paradigm for information processing, drawing inspiration from the intricate workings of biological nervous systems, most notably, the human brain's remarkable capacity to process and interpret complex data. Figuratively speaking, ANNs emulate the neural network within the human brain, as depicted in Figure 1, to undertake intricate computational tasks with remarkable efficiency and adaptability.

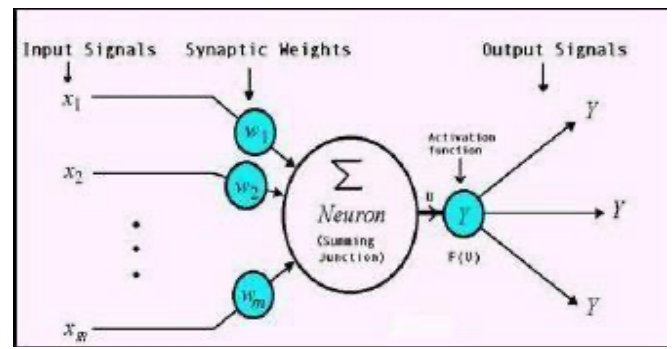

$$u = \sum_{j=1}^{m} W_j X_j \qquad Y = F(u)$$

Fig 2. Basic Neural Network model

In essence, ANNs replicate the fundamental structure of biological neural networks. These networks consist of neurons interconnected by synapses, mirroring the communication pathways within the human brain. In this neural framework, synapses serve as sensors, adept at capturing inputs from the surrounding environment. Meanwhile, the soma, or the central body of the neuron, stands as the fundamental processing unit, orchestrating the intricate web of computations that define the neural network's functionality. This intricate interplay of synapses and soma embodies the essence of ANNs, as they leverage this biologically inspired architecture to facilitate complex data analysis and pattern recognition.

To better appreciate the conceptualization of ANNs, refer to the illustrative representation depicted in Figure 2.

This simplified model elucidates the core structure that underpins ANNs' functionality, offering a visual framework for understanding their modus operandi. Through the deployment of ANNs, we endeavor to harness the inherent capabilities of neural networks in accelerating the process of knowledge extraction and data interpretation, revolutionizing diverse domains, including medical diagnostics, where precise and rapid decision-making is of paramount importance.

In the realm of Artificial Neural Networks (ANNs), the journey of information processing embarks as neurons engage with a set of inputs. These inputs are not merely processed in isolation; rather, they undergo a transformative voyage orchestrated by the activation function, ultimately yielding corresponding outputs. Central to this process are the weights associated with each connection, a critical determinant of both the strength and sign of the interactions within the neural network. Tailoring the number of output layers in the network is a pivotal design consideration, and the choice of the activation function intricately influences the nature of the produced outputs.

Within the intricate tapestry of ANNs, two primary categories of neural network structures emerge: acyclic or feed-forward networks and cyclic or recurrent networks. The former, exemplified by the feed-forward network, operates as a function of its current inputs, devoid of any internal states beyond the weight coefficients themselves. In stark contrast, recurrent networks take a more intricate approach by looping their outputs back into their own inputs, thereby endowing the system with memory-like capabilities. This internal feedback loop allows recurrent networks to exhibit dynamic behaviors, such as convergence to a stable state, oscillations, and in some instances, even chaotic patterns. Furthermore, the network's response to a given input is intricately linked to its initial state, often shaped by previous inputs, thus bestowing upon it the capability to support short-term memory, a feature that finds significant utility in various computational contexts.

The architectural design of ANNs can encompass a spectrum of complexity, ranging from a single hidden layer to networks replete with multiple hidden layers. These hidden layers operate in tandem to process the incoming inputs, unraveling intricate patterns and relationships embedded within the data. The collective activation levels across the network form a dynamic system, with the potential to converge to a stable equilibrium, oscillate in a rhythmic fashion, or exhibit chaotic behaviors, depending on the intricacies of the network's structure and the inputs encountered.

In the following sections of this paper, we delve deeper into the operational dynamics of these neural network structures, exploring their capacity to model complex systems, adapt to varying data distributions, and, most importantly, advance our understanding of how ANNs can be harnessed to revolutionize the landscape of prostate cancer detection and diagnosis. Through a comprehensive exploration of these concepts, we aim to provide a solid foundation for the application of ANNs in medical diagnostics, elucidating their potential to expedite early detection, enhance accuracy, and ultimately improve patient outcomes in the context of prostate cancer.

Consider the simple network shown in Fig 3 which has two inputs, two hidden units and an output unit. Given an input vector $X = (x_1, x_2)$, the activation of the input units is set to $(a_1, a_2) = (x_1, x_2)$, and the network computes

$$a_5 = g(W_{3,5} a_3 + W_{4,5} a_4)$$
$$= g(W_{3,5} g(W_{1,3} a_1 + W_{2,5} a_2) +$$
$$+ W_{4,5} g(W_{1,4} a_1 + W_{2,4} a_2)) \quad \ldots \text{Eqn.1}$$

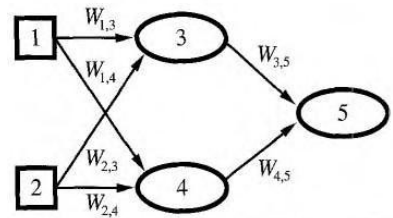

Fig. 3 A simple neural network with two inputs and two hidden units and a single output.

Thus, by expressing the output of each hidden unit as a function of its inputs, we have represented $a_5$ as a function of the network inputs along with the weights. There exist simple single layer feed forward neural networks, but we have used the multi layered feed-forward net.

### B. Multilayer Feed Forward Neural Networks: Unlocking Complexity & Versatility

When we delve into the domain of multilayer feed-forward neural networks, we uncover a realm of computational sophistication characterized by the presence of multiple hidden layers within the model. This multi-layered architectural configuration bestows upon the network a host of advantages, chief among them being an augmented capacity for complex computations and the ability to accommodate a broader spectrum of hypotheses, thereby enhancing its representational power. Each hidden layer within this multifaceted model serves as an embodiment of a threshold function, poised to evaluate and process the inputs received from the network's input layer.

At the heart of this intricate neural network framework lies the aggregation of inputs, a critical precursor to their transformation through the transfer function denoted as "f." It is through this transfer function that the neural network refines and structures the incoming information, ultimately producing meaningful outputs. In the vast lexicon of neural network architectures, a diverse array of threshold functions finds application, each tailored to specific computational requirements and analytical contexts. These threshold functions, encapsulating distinct mathematical properties and behaviors, offer a rich tapestry of options for fine-tuning the neural network's operations to align with the task at hand.

One of the pivotal threshold functions employed in neural networks is the "hard-limit transfer function." Aptly named, this function exerts a stringent control over the neuron's output, constraining it to one of two discrete values: 0 or an alternative output value contingent upon the aggregate input arguments supplied by the network's input layer. This binary nature of the hard-limit transfer function makes it particularly well-suited for scenarios where decisions or classifications are dichotomous, exemplifying its utility in various computational and analytical contexts.

The sum of inputs acts as the parameters to the transfer function f. The various threshold functions used in the neural networks [2] are:

**i)**       **Threshold Function**

The threshold function also called as the hard-limit transfer function limits the output of the neuron to either 0, if total inputs arguments from the input layer have a value less than 0 or 1 if the net value is greater than or equal to 0.

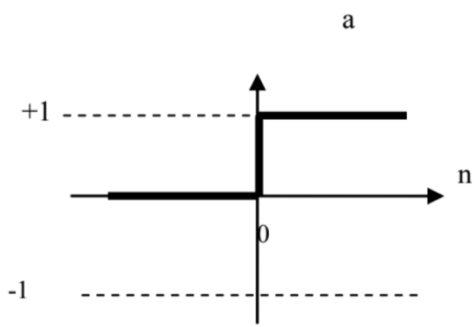

Fig. 4 Threshold function

**i)**       **Linear Transfer Function**

The linear transfer function is used to differentiate between the net input value lying on one side of the linear line through the origin.

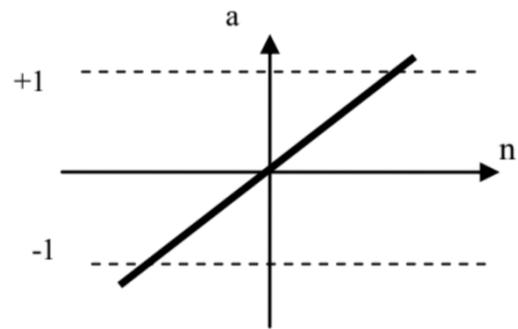

Fig. 4 Linear Transfer function

**i)**       **Log – Sigmoid Transfer Function**

The log-sigmoid transfer function is most used in back-propagation models as it is differentiable.

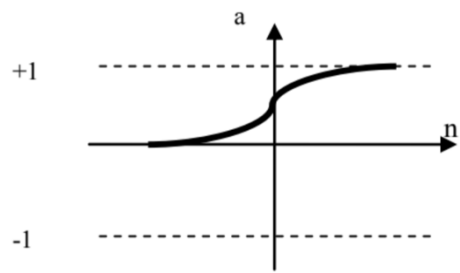

Fig. 4 Log Sigmoid Transfer function

With more hidden units in the model, and the hypotheses space increasing, the back propagation model helps us to train the model more efficiently. In the back propagation model the output obtained from the training is compared with actual outputs and we calculate the error.

## III. LINKING ANN's TO PROSTATE CANCER ANALYSIS

### A. Data Analysis and Model Description

The existing diagnostic methods for prostate cancer frequently yield a considerable number of False Positive Test Results (FPTRs), posing significant challenges in terms of precision and accuracy. To address this pressing concern, we turn to the formidable capabilities of Artificial Neural Networks (ANNs) to construct models that not only curtail the incidence of FPTRs but also endeavor to heighten the overall diagnostic accuracy. Our foundational dataset comprises a comprehensive cohort of 1983 patients [3], hailing from diverse backgrounds and medical histories. Within this cohort, 1551 patients were conclusively diagnosed with prostate cancer, while 432 individuals, following a battery of initial tests and biopsies, were subsequently deemed free from prostate cancer. This invaluable dataset was thoughtfully sourced from Weil Medical College, Department of Urology, and has been meticulously curated with unwavering guidance from a cadre of dedicated medical professionals, including doctors and physicians. To safeguard the privacy and confidentiality of the patients, all personally identifiable information, including names, was rigorously omitted from our analysis.

The immense dataset at our disposal was thoughtfully subdivided into four distinct smaller datasets, each serving as a microcosm of the broader patient population. The first dataset encompassed 400 patients diagnosed with prostate cancer and an additional 100 who were conclusively free from the disease. This division was mirrored in the subsequent two datasets. In the final dataset, 351 patients received a prostate cancer diagnosis, while 132 individuals emerged unscathed from the specter of cancer. This strategic segmentation allowed us to train the neural network model individually on each of these datasets, thereby facilitating a granular and contextually nuanced understanding of the model's performance across diverse patient groups.

The neural network model we employed adheres to a structured architecture, commencing with an input layer capable of accommodating a multitude of inputs denoted as x1, x2, x3, and so forth. Each of these inputs is assigned a specific weight, represented by w1, w2, w3, and so forth, reflecting the nuanced importance and impact of individual input features. The hallmark of this model lies in its ability to consolidate these inputs, effectively summing the products of inputs and their respective weights for all variables, thus generating an aggregate signal that is then conveyed to the subsequent neural layers for further processing. For a visual representation of this data flow, please refer to Figure 2, which provides a schematic elucidation of this crucial processA.

$$u = \sum_{j=1}^{m} wjxj \quad \ldots \text{Eqn 2}$$

We have used the multi - layered feed forward model which would back propagate the outputs back to the hidden units i.e., we back propagate the model. Input vectors and the corresponding target (output) vectors are used to train the network until it can approximate a function; associate input vectors with specific output vectors or classify the input vectors in an appropriate way as specified by the model. In the standard back propagation models the network weights are moved along the negative of the gradient of the performance function. The term back propagation refers to the way the gradient is computed for nonlinear multilayer networks. In the more specific back propagation models, they tend to provide more accurate results as expected from the targets. A new input in this model will lead to an output like the correct output for input vectors used in training that are similar to the new input being presented [4].

The dataset consists of four variables which are selected and have been of much relevance in diagnosing the cancer among men [5]. The four variables used are as:

- Age of the patient
- Prostate size in grams (using Ultrasound imaging)
- Most Recent PSA Level
- Most Recent free PSA Level

These four variables were selected and distributed into the data sets. These variables were obtained from the patients diagnosed in Weill Medical College. These inputs are fed to the networks in the form of input vectors.

We need to distribute the data sets into 3 different subsets where 60% of the input vectors are used to train the network, 20% of the input vectors are used to validate the model i.e., the network that has been created in this case and the remaining 20% of the input vectors are used to rest the network for the generalization. This combination of the training, validating and testing data can be configured but after a series of tests we did find this to give better results as compared to other combinations. This percentage distribution is also dependent on the amount of input vectors that we have and the number of input variables that the network has.

### B. Styles of Training

In the intricate domain of neural network training, the choice of training method holds paramount significance, as it fundamentally dictates how the model adapts and evolves in response to data. In our research endeavor, we have primarily embraced the batch training method, a methodological approach that introduces distinct characteristics and advantages into the training process. Under the purview of batch training, a critical facet manifests—the weight and bias adjustments take place solely upon the comprehensive presentation of an entire batch of input vectors, coupled with their corresponding target values, to the neural network. This approach engenders a synchronized framework for weight updates, treating the individual inputs as if they were concurrently processed, despite their sequential arrangement within a data structure, such as an array. This training cycle perseveres until specific pre-specified conditions have been met or until the model has successfully attained its predefined objectives, encapsulating the essence of batch training.

To elucidate this process further, it is imperative to underscore the central role played by target vectors, judiciously defined during the network's configuration phase. These target vectors serve as benchmarks against which the model's generated outputs are meticulously compared and assessed. Furthermore, the process commences with the initialization of the model's weights, thoughtfully considering all input vectors, thus laying the foundational groundwork for the impending training iterations. As the training journey unfolds, it sets in motion a profound learning process, conceptually framed as an optimization quest within the expansive weight space.

The crux of this learning voyage hinges on the classical metric of error computation, wherein the model's generated outputs are subjected to rigorous evaluation against the authentic target values. It is the magnitude of this error that serves as the lodestar for subsequent weight adjustments. Based on the magnitude of this error, the neural network fine-tunes its internal weights, and the model is further refined. This iterative process of backpropagation permeates through the intricate layers of the neural network, facilitating a continuous refinement of its configuration and structure.

The set of outputs generated from the model will be compared to the target vectors which had been specified earlier to the network. We need to initialize the weights to all the input vectors. The training is started for the input vectors and the net starts to learn. The learning is formulated as an optimization search in weight space. We have used the classical measure of error between the outputs that are obtained from the network that has been configured and the actual target values. Depending upon this error values, the weights would change we would train the network again as this would be back propagated to the hidden layers in the model. The squared error for a single transition of input vector would be:

E = 0.5 Err$^2$
 = 0.5 (Output – network output)      … Eqn 3

We can use the gradient descent to reduce the squared error by calculating the partial derivative of E with respect to each of the weight. We have,

$\frac{\partial y}{\partial x}$ = - Err * $\frac{\partial}{\partial w_j}$ (Output – Activationfn($\sum_{j=0}^{m} Wjxj$))

… Eqn 4

For the sigmoid function we have the derivative as:

g` = g(1-g)

So from the above equation 4 we can further modify it as :

 = - Err * g`(inputs) * xj            … Eqn 5

So the corresponding weight would be updated as follows:

W$_j$ ← W$_j$ + α* Err * g`(inputs) * x$_j$      … Eqn 6

where α is the learning rate. The learning rate α is multiplied times the negative of the gradient to determine the changes to the weights and the biases fed to the network. The larger the learning rate, the bigger the step. It the learning rate is made to large, the algorithm becomes unstable which has been proved in this project. We have shown the training of the model for different values of the learning rate and how the model behaves in case of the changes in the learning rate and how it affects the values of weights and biases in the iterative back propagation process. If the error is positive, then the network output is too small and so the weights are increased for the positive inputs and decreased for the negative inputs. The opposite happens when the error is negative [Rosenblatt (1957)].

The model trains the input vectors, adjusts the weights as per the algorithm to reduce the error. Each iteration is called as an epoch. Epochs are continued till the goal is reached. The goal can be state in the form of either the input vectors reaching the goal state i.e., the target values which have been fed to the network or some specific value mentioned with respect to the descending gradient. Now the number of epochs keeps on changing as we train the model several times. The epochs represent the number of times the input vectors are iterated over to be trained so that the mean squared error between the inputs and the specified outputs is as small as possible. Epochs can also be stopped by providing some pre- specified condition like the threshold goal state value falling below a certain level or it is having a very high value as compared to the target values. The conditions could be the weight changes are too small. In other methods we calculate the gradient for the whole training set by adding up all the gradient contributions from the Eqn. 6 before we update the weights again and feed to the network for other iteration.

For the back propagation model, we back- propagate the error from the output layer to the hidden layers. The back propagation process emerges directly from a derivation of the overall error gradient. At the output layer the weight update would be based on Eqn. 6. As we have multiple outputs mentioned in this case we would have the modified version of the Eqn. 6 as:

W$_{j,i}$ ← W$_{j,i}$ + α* Err$_i$ * g`(inputs) * x$_j$      … Eqn 7

The i in the equation represents to the ith component of the error vector. The modified weights need to be sent to the hidden layers now with new values. This clearly describes that the error back propagation concept. This lists that some hidden node in any of the hidden layers would be responsible for the error that occurs in the run. So now the error that we come across between the target vectors and the output values obtained from the run would be back propagated. The modified error values would be divided according to the strength of the connection between the hidden node and the output node and are propagated back to the set of hidden layers. The propagation rule can be mentioned as:

Єj = g` (input j) $\sum_{j=1}^{m} Wji$ Єj      …Eqn 8

We can list the basic steps involved in back propagation model as:

- Compute the values Єj for the output units, using the observed error.
- Starting with output layer, repeat the following two steps for each layer in the network until the earliest hidden layer is reached:
  - Propagate the Єj values back to the previous layer in the model.
  - Update the weights between two layers to move towards the goal state values.

## IV. BUILDING THE NETWORK AND TRAINING

### A. Dataset Distribution

From the four data sets of 1983 patients we build the network model using the Matlab tool for neural networks. We trained the network with different training functions and different learning rated over a multiple number of times.

We need to present the input vector with four variables and in set of around 500 values as:

P = [54, 52; 12 14; 1.2 2.3; 11 18];
T = [1 10]

In the input vector we have the first input vector consisting of four input variables in the form of:

P1 = 54 12 1.2 11
P2 = 52 14 2.3 18

And their corresponding target values as 1 and respectively indicating that the first patient p1 in the data set was diagnosed with prostate cancer and the patient p2 was not diagnosed with prostate cancer. Next the network is built, we started with a single hidden layer, and after many tests we checked, the accuracy gained as compared to two hidden layers was less.

### B. Training Functions

We trained the model over a list of training functions and the various results and observations are as follows:

i. Batch Gradient Descent
ii. Variable Learning Rate
iii. Resilient Backpropagation
iv. Quasi Newton Algorithm
v. Levenberg – Marquardt Algorithm

#### i. Batch Gradient Descent

The batch steepest descent training function is traingd. The weights and biases are updated in the direction of the negative gradient of the performance function. We would stop the training of the performance function specified falls below the goal mentioned, if the magnitude of the gradient is less than value specified, or if the training time is longer than time seconds. After training the network we simulate the network to check for the output values after the model has been trained n number of times.

For the first data set we have in the graph for X – axes as the number of inputs and Y- axis as the squared error-Performance:

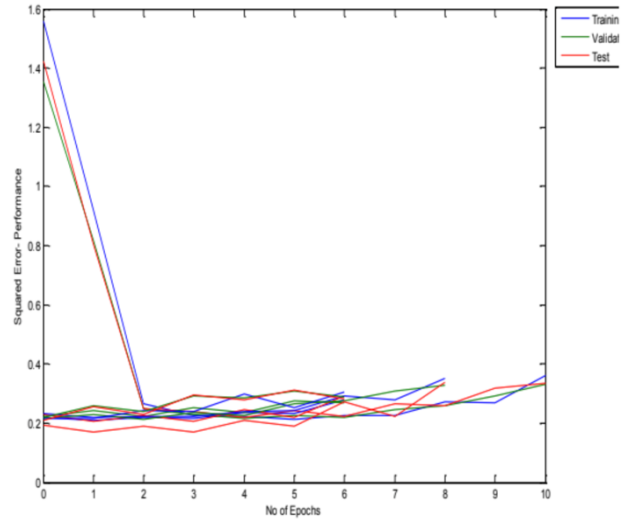

Fig. 7 traingd No of Epochs vs Squared Error Performance

For four different runs of learning rate=0.07 we check that the each run has different number of epochs and the squared errors decreases till it reaches 0. Then we simulate the network response to the original set of inputs to check for the values as compared to target values and we can check that as follows:

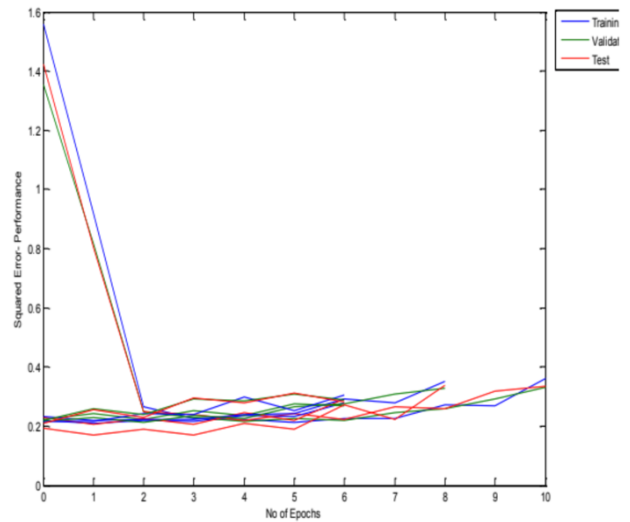

Fig. 8 traingd Input Vectors *p* vs Neural Network values

We check for the accuracy of the input vectors and we can see that most of the values are reaching the upper surface of value 1 with some exceptions for value approaching 0.5 depicting the patients not diagnosed with prostate cancer. Then we tested the network for the accuracy in this case for two input vectors q and r.

```
q =[41 62 72 60 75 70 79 71 52 54
;21 0 44 32 61 32.4 0 0 72.7 65.5
;3.3 0 4.2 7.3 10 5.2 0 0 5 6.7 ;11
0 26 8 17 8 0 0 20 11];
b=sim(net,q)
```

```
r =[66 68 36 65 53 55 65 72 62 70 56;
59.1 76.6 14.4 49 22 40 0 69 117 67.4
39; 1.8 1.8 0.2 7.5 0 4.2 0 8.9 17.9
5.4 6.3; 51 31 0 11 0 20 0 19 22.3 26
10];
c=sim(net, r)
```

We check for the output values for the input vector q and r as follows:

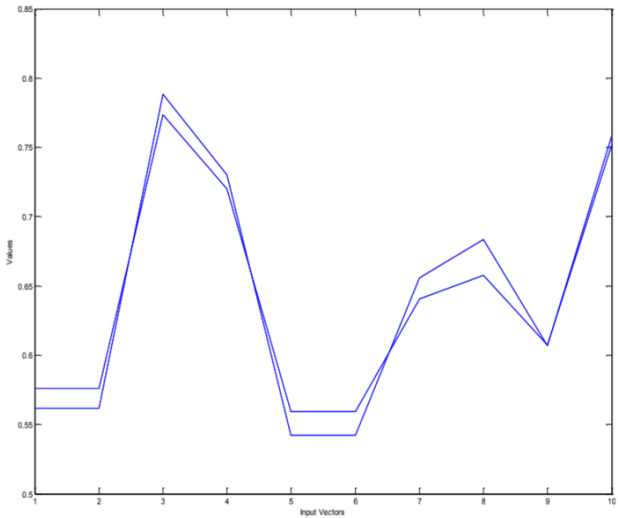

Fig. 9 traingd Input Vectors *q* vs Neural Network values

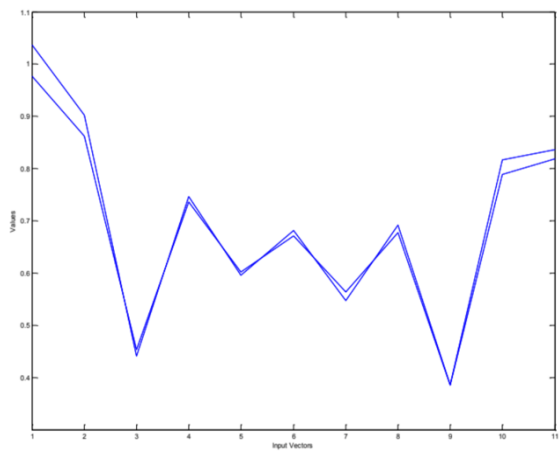

Fig. 10 traingd Input Vectors *r* vs Neural Network values

The training algorithm was too slow as it took more number of epochs to reach the goal state or reaching the condition for stopping the training.

### ii. Variable Learning Rate

We have two variable learning rate algorithms [4]. The performance of the algorithm is very sensitive to the proper setting of the learning rate. If the learning rate is set too high, the algorithm can oscillate and become unstable. If the learning rate is too small, the algorithm takes too long to converge. It is not practical to determine the optimal setting for the learning rate before training, and, in fact, the optimal learning rate changes during the training process, as the algorithm moves across the performance surface. We need to allow the learning rate to change during the training process. An adaptive learning rate attempts to keep the learning step size as large as possible while keeping learning stable. The learning rate is made responsive to the complexity of the local error surface. An adaptive learning rate requires some changes in the training procedure used by the previous method. First, the initial network output and error are calculated. At each epoch new weights and biases are calculated using the current learning rate. New outputs and errors are then calculated. This procedure increases the learning rate, but only to the extent that the network can learn without large error increases. Thus, a near-optimal learning rate is obtained for the local terrain. When a larger learning rate could result in stable learning, the learning rate is increased. When the learning rate is too high to guarantee a decrease in error, it is decreased until stable learning resumes. We check for these heuristic techniques, which were developed from an analysis of the performance of the standard steepest descent algorithm.

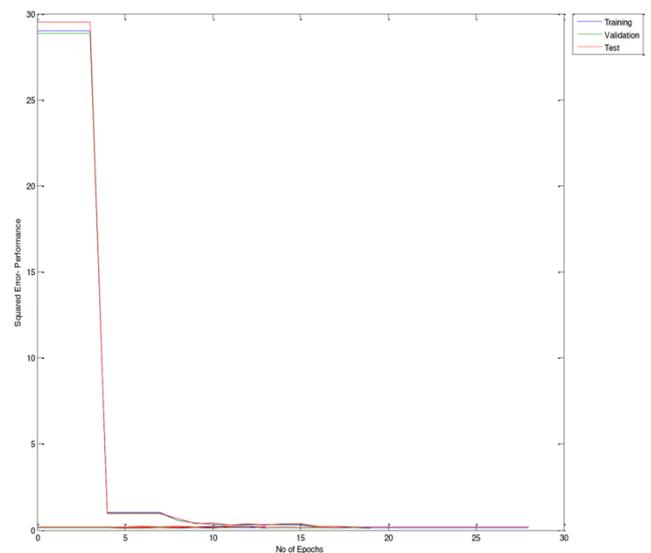

Fig. 11 traingda No of Epochs Vs Squared Error Performance

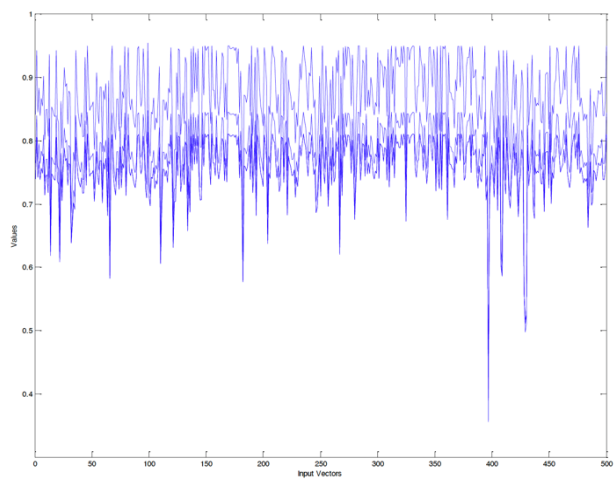

Fig. 12 traingda Input Vectors p vs Neural Network values

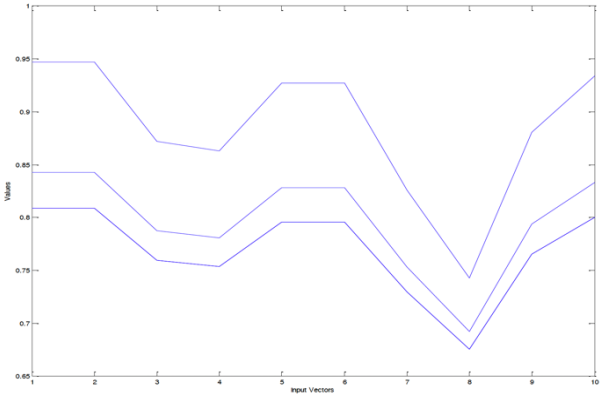

Fig. 13 traingda  Input Vectors q vs Neural Network values

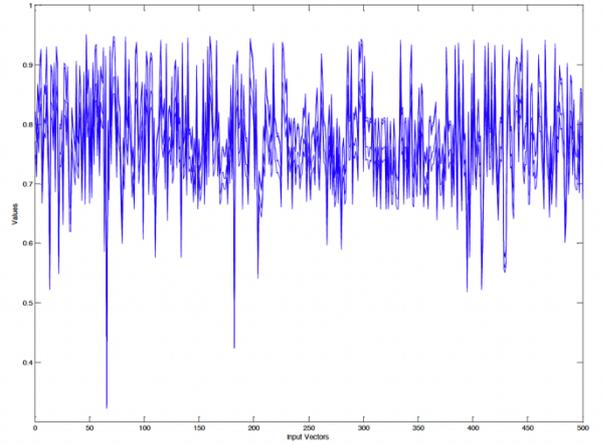

Fig. 16 traingdx Input Vectors p vs Neural Network values

In this case the function traingdx combines adaptive learning rate with momentum training. It is invoked in the same way as traingda, except that it has the momentum coefficient mc as an additional training parameter.

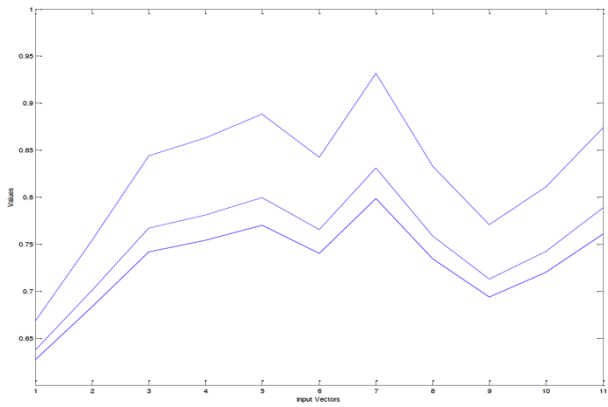

Fig. 14 traingda Input Vectors r vs Neural Network values

And for the function traingdx we have:

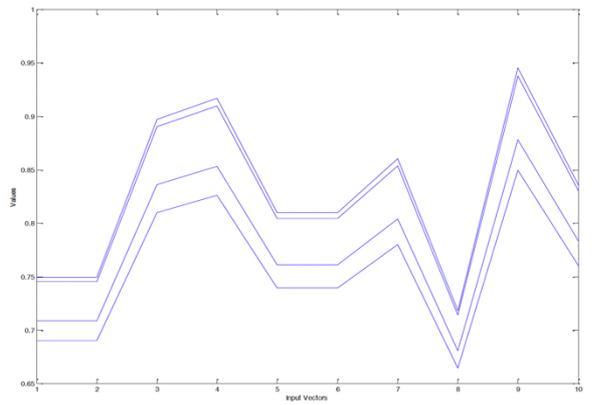

Fig. 17 traingdx Input Vectors q vs Neural Network values

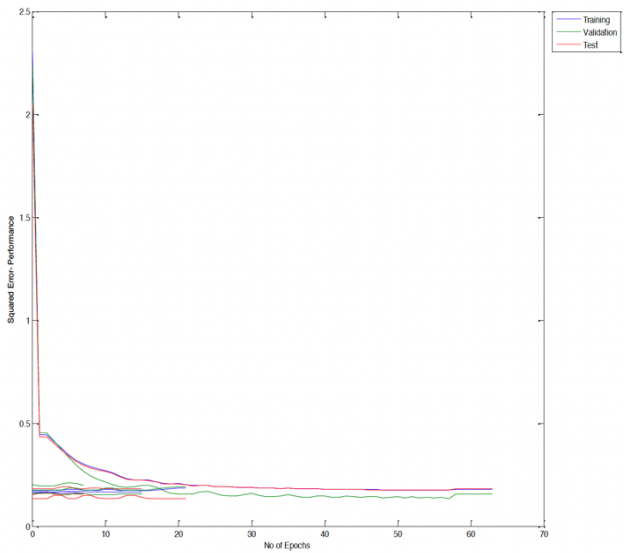

Fig. 15 traingdx  No of Epochs vs Squared Error Performance

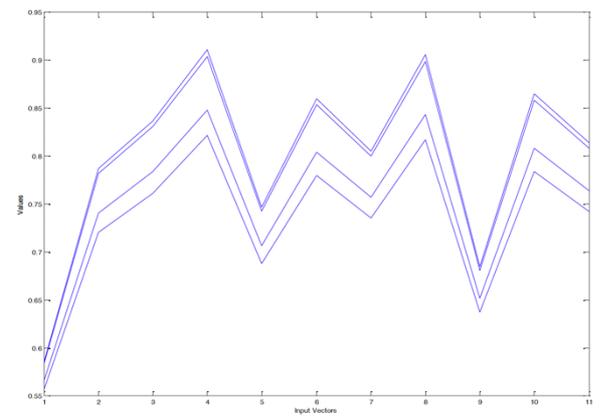

Fig. 18 traingdx Input Vectors r vs Neural Network values

### iii. Resilient Back Propagation

Multilayer networks typically use sigmoid transfer functions in the hidden layers. These functions are often called "squashing" functions, because they compress an infinite input range into a finite output range. Sigmoid functions are characterized by the fact that their slopes must approach zero as the input gets large. This causes a problem when you use steepest descent to train a multilayer network with sigmoid functions, because the gradient can have a very small magnitude and, therefore, cause small changes in the weights and biases, even though the weights and biases are far from their optimal values. Only the sign of the derivative is used to determine the direction of the weight update; the magnitude of the derivative has no effect on the weight update. The size of the weight change is determined by a separate update value. The update value for each weight and bias is increased by a factor whenever the derivative of the performance function with respect to that weight has the same sign for two successive iterations. The update value is decreased by a factor whenever the derivative with respect to that weight changes sign from the previous iteration. If the derivative is zero, then the update value remains the same. Whenever the weights are oscillating, the weight change is reduced. If the weight continues to change in the same direction for several iterations, then the magnitude of the weight change increases.

This training function is generally much faster than the standard steepest descent algorithm. It also has the nice property that it requires only a modest increase in memory requirements. You do need to store the update values for each weight and bias, which is equivalent to storage of the gradient.

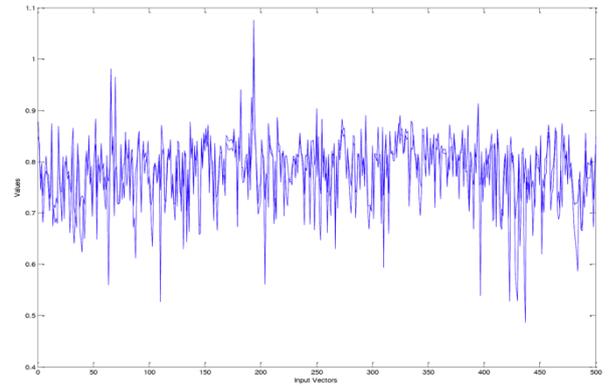

Fig. 20 trainrp Input Vectors p vs Neural Network values

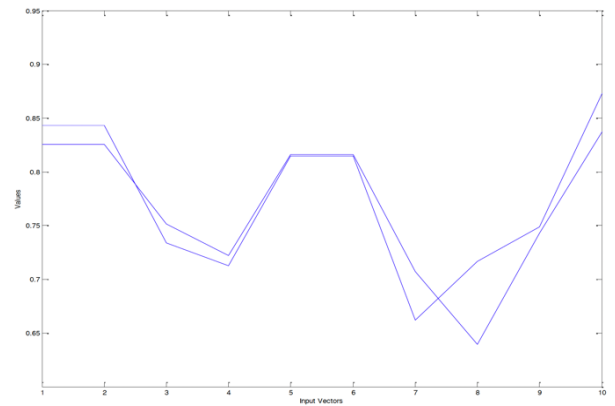

Fig. 21 trainrp Input Vectors q vs Neural Network values

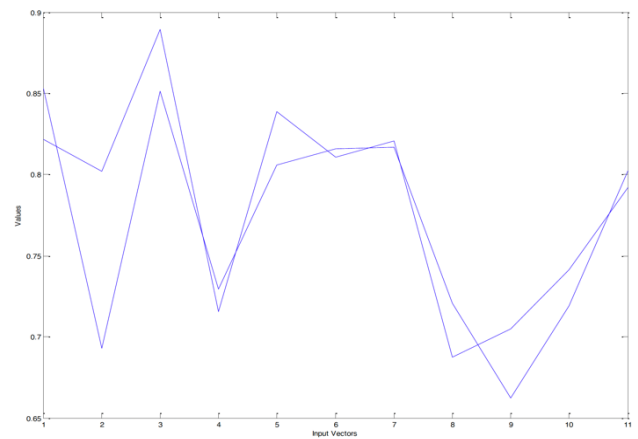

Fig. 22 trainrp Input Vectors r vs Neural Network values

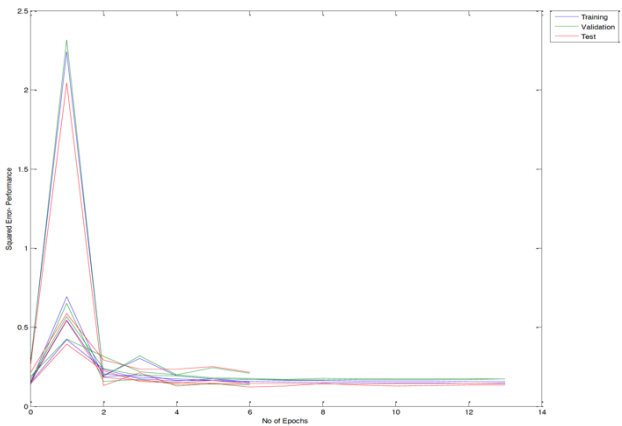

Fig. 19 trainrp No of Epochs vs Squared Error Performance

### iv. Quasi Newton Algorithms

Newton's method is an alternative to the conjugate gradient methods for fast optimization. Newton's method often converges faster than conjugate gradient methods. Unfortunately, it is complex and expensive to compute the Hessian matrix for feed forward neural networks. There is a class of algorithms that is based on Newton's method, but which doesn't require calculation of second derivatives. These are called quasi-Newton (or secant) methods. They update an approximate Hessian matrix at each iteration of the algorithm. The update is computed as a function of the gradient. The quasi-Newton method that has been most successful in published studies is the Broyden, Fletcher, Goldfarb, and Shanno (BFGS) update.

We have the equation as:

This algorithm requires more computation in each iteration and more storage than the conjugate gradient methods, although it generally converges in fewer iterations. The approximate Hessian must be stored, and its dimension is n x n, where n is equal to the number of weights and biases in the network. For very large networks it might be better to use Rprop or one of the conjugate gradient algorithms. For smaller networks, however, trainbfg can be an efficient training function

### v. Levenberg - Marquardt Algorithms

The Levenberg-Marquardt algorithm was designed to approach second-order training speed without having to compute the Hessian matrix. This is the faster than the other methods considered; also the accuracy level as compared to the other models is high. Due to non-storage of the matrix values the processing speed is reduced and we can obtain the results much faster. Instead we use a Jacobian matix $J_T$ that contains the first derivatives of the network errors with respect to the weights and the biases. We did check that for smaller set of input vectors in hundreds the training function performs very well but if the number of input vectors increases drastically then the training function performance decreases and the time taken to complete the training with respect to the number of epochs is huge.

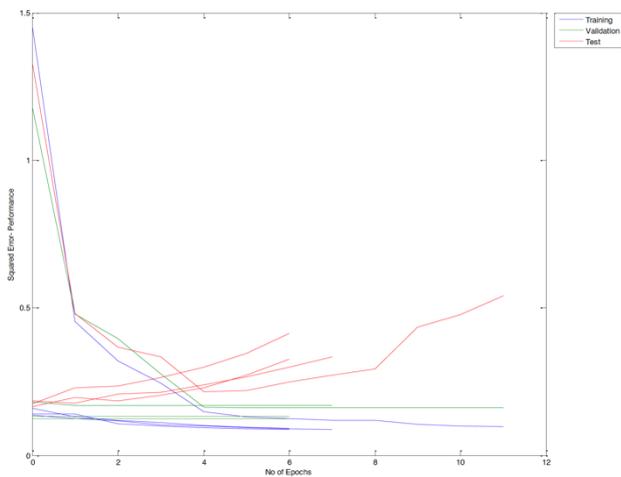

Fig. 23 trainlm No of Epochs vs Squared Error Performance

We can check from the graph that the number of epochs taken in this case is leas and it almost reduces the mean squared errors thus giving a better training function to build the model. As for the input vectors and vectors q and r we have the output values as:

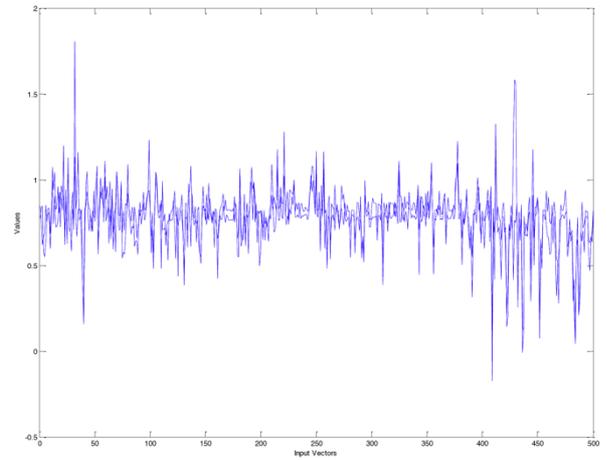

Fig. 24 trainlm Input Vectors p vs Neural Network values

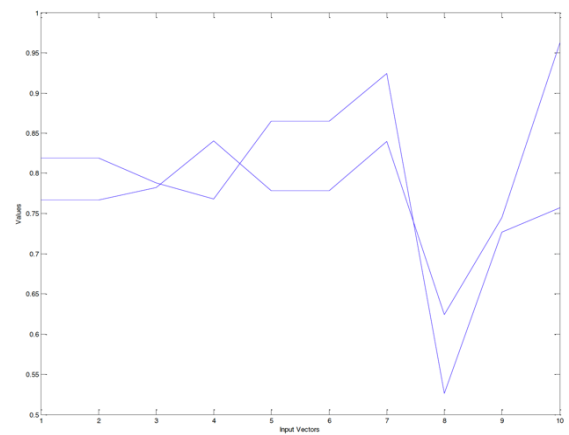

Fig. 25 trainlm Input Vectors q vs Neural Network values

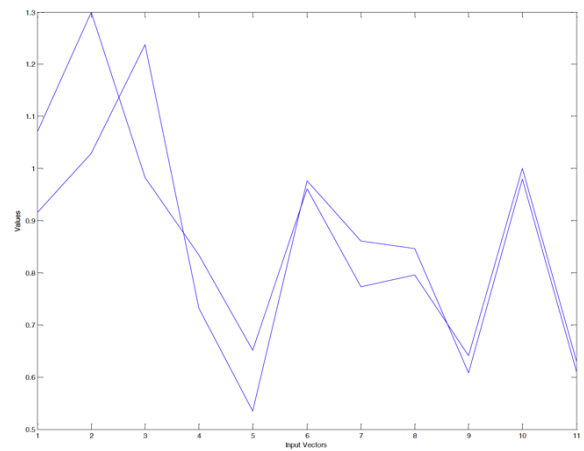

Fig. 26 trainlm Input Vectors r vs Neural Network values

## V. COMPARISONs AMONG THE TRAINING ALGORITHMS:

Various training algorithms are dependent upon the training data and the model. The factors affecting this could be the complexity of the problem, number of input vectors, hidden layers, error goal and whether the network is being used for pattern recognition or function approximation. We can check from the above runs for all the four different data sets, it tends to have different accuracy percentage and the number of epochs taken is less. The comparisons of ANNCRIPS with respect to other non-neural networks models is shown in table 1.1 and comparisons between ANNCRIPS and other neural networks models is shown in table 1.2.

Table 1.1

| Reference | Study Cohort | Predictive Model | Precision (%) |
|---|---|---|---|
| ANNCRIPS | Pre-screening of prostate cancer | MLP – Back Propagation model | 81 |
| Optenberg et al.[8] | Suspicion for prostate cancer | Mulitple Logisitc regression | 81 |
| Benecchi[9] | Urologic symptoms, abnormal PSA or DRE | Neuro-fuzzy inference system model | 80 |
| Herman et al.[10] | Screening cohort | Look-up table | 62-68 |

Table 1.2

| Reference | Training Cohort | Input variables | Precision(%) |
|---|---|---|---|
| ANNCRIPS | Screening population | Age PSA tPSA fPSA | 81 |
| Porter et al.[11] | Screning population | Age PSA Prostate volume PSAD DRE TRUS | 77 |
| Stephan et al.[12] | Mixed screened & non-screened, PSA 2-10 ng/mL | Age PSA %fPSA Prostate volume DRE | 65-93 |
| Stephan et al.[13] | Mixed screened & non-screened PSA 4-10 ng/ML | Age PSA %fPSA Prostate volume DRE | 75-85 |
| Matsui et al.[14] | Japense screening population or urologic symptoms, PSA 2-10 ng/mL | Age PSA %fPSA Prostate volume TZ volume PSAD PSA-TZ DRE LUTS | 79 |
| Finne et al.[15] | Screening population aged 55-67 years, PSA 4-10 ug/mL | PSA %fPSA Prostate volume DRE Age | 76 |

## VI. FURTHER IMPROVEMENTS:

We could test the model over a huge data set and train it multiple numbers of times with all the training functions. The inclusion of more variables would enhance the accuracy level and can help us to predict the occurrence of prostate cancer among the patients. As the number of variables increases, we get more number of input vectors over which we can train the model. Also, by changing the number if hidden layers we can accommodate a large number of hypotheses to train the model.

## VII. CONCLUSION:

Thus, we see that Artificial Neural networks can be efficiently used to diagnose cancer at an early stage enabling us to reduce the number of false positive test results. Also, this learning technique takes into consideration a large number of factors like the different input arguments from the patients, the number of hidden layers in the network, etc. After training the model over a huge data set and simulating the model by validating over the set of input data present we could check the remaining test set for its accuracy.

## VIII. ACKNOWLEDGEMENTS:

1. Prof. Bart Selman. Cornell University
2. Dr. Ashutosh Tewari, Cornell Weill Medical College, Dept. of Urology
3. Douglas S. Scherr, M.D., Cornell Weill Medical College, Dept. of Urology
4. Micheal Herman, Cornell Weill Medical College, Dept. of Urology

5. Robert Leung, Cornell Weill Medical College, Dept. of Urology
6. Karan Kamdar, University of Mumbai
7. Late. Prof. K.G. Balakrishnan, University of Mumbai